\documentclass[acmsmall,nonacm]{acmart}
\usepackage{makecell}
\usepackage{subfig}
\usepackage{multirow}
\usepackage{cleveref}
\usepackage{xcolor}
\crefname{figure}{Fig.}{Figs.}
\crefname{table}{Table.}{Tables.}
\AtBeginDocument{%
  }

\renewcommand\footnotetextcopyrightpermission[1]{}

\acmJournal{TOMM}

\begin{document}

\title{Diff-VF: Training-free High-quality Long Video Generation via Diffusion Model}

\author{Haoning Yang}
\email{yanghaoning@sjtu.edu.cn}
\orcid{0009-0008-7100-7687}
\affiliation{%
  \institution{Shanghai Jiao Tong University}
  \city{Shanghai}
  \country{China}
}

\author{Xinyuan Chen}
\authornote{Corresponding authors.}
\orcid{0000-0002-5517-7255}
\affiliation{%
  \institution{Shanghai Artificial Intelligence Laboratory}
  \city{Shanghai}
  \country{China}
}
\email{chenxinyuan@pjlab.org.cn}

\author{Yaohui Wang}
\orcid{0009-0002-9487-6187}
\affiliation{%
  \institution{Shanghai Artificial Intelligence Laboratory}
  \city{Shanghai}
  \country{China}
}
\email{wangyaohui@pjlab.org.cn}

\author{Guo Lu}
\authornotemark[1]
\orcid{0000-0001-6951-0090}
\affiliation{%
  \institution{Shanghai Jiao Tong University}
  \city{Shanghai}
  \country{China}
}
\email{luguo2014@sjtu.edu.cn}

\renewcommand{\shortauthors}{Yang et al.}

\begin{abstract}
  Recently, diffusion models have made great progress in video generation. However, most existing video diffusion models are trained with short videos, and degrade when extrapolated to long videos, struggling to maintain long-range temporal coherence while retaining diverse motions. To generate consistent, high-quality and dynamic long videos, we propose Diff-VF, a training-free, plug-and-play and model-agnostic framework that converts existing short-video diffusion backbones into long-video generators without modifying or fine-tuning the base model. Diff-VF couples three complementary strategies: Hybrid Noise Initialization (HNI) to constrain global semantics, Weighted Window Sampling (WWS) to remove inter-window discontinuities, and Temporal Extended Sampling (TES) to establish long-range dependencies with a timestep-varying fusion. We further extend Diff-VF to long-video enhancement via Skip Residual Guidance that balances fidelity and realism through timestep-dependent guidance. VBench-Long evaluation results show that Diff-VF achieves a more favorable balance between temporal coherence and motion diversity than base models and recent training-free long video generation baselines, including FreeNoise, FreeLong, and RIFLEx, while maintaining competitive frame-wise quality. Experiments on two base models demonstrate the applicability to video diffusion models with different spatial-temporal modeling strategies. Extensive ablations validate the contribution of each component and hyperparameters.
\end{abstract}





\begin{teaserfigure}
    \centering
    \includegraphics[width=0.99\linewidth]{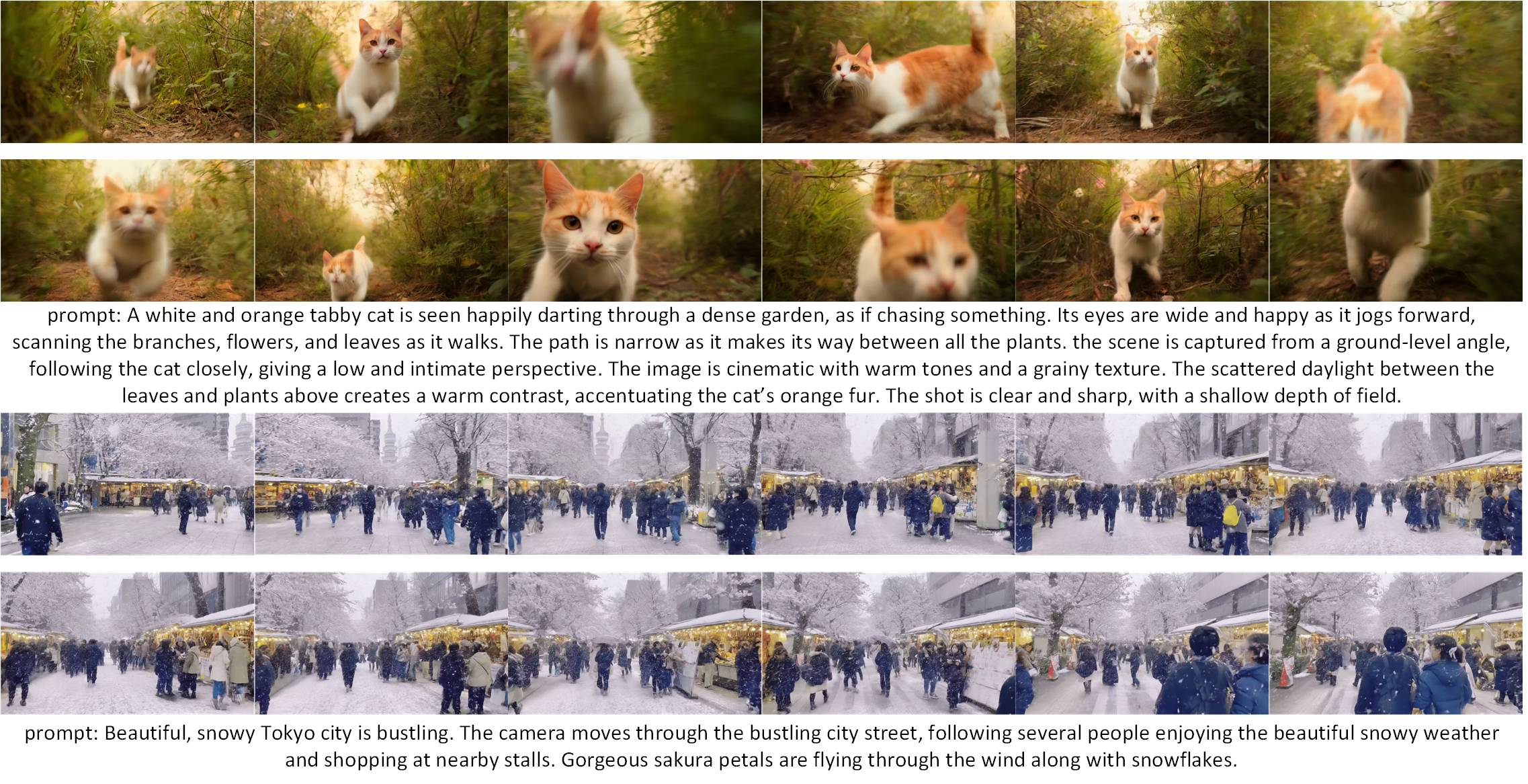}
    \caption{Illustration of long videos generated by our method based on HunyuanVideo. Both of them are $4\times$ the original length of videos generated by the base model. 
    }
    \label{teaser}
\end{teaserfigure}

\maketitle

\section{Introduction}
\label{intro}
Diffusion models allow for a dramatic improvement in the quality of image generation \cite{ddpm,ddim,ldm,beats,cooking}, greatly exceeding previous state-of-the-art methods, such as VAEs \cite{vae} and GANs \cite{gan}. Based on the success in image domain, a considerable number of researches have sought to extend the application of diffusion models to video domain \cite{vdm, lvdm,lavie,videocrafter1,videocrafter2,i2vgen,animatediff,sora,cogvideo,cogvideox,hunyuanvideo,wan}. Furthermore, the strong generative power of diffusion models can be leveraged to enhance the quality of videos \cite{seine,upscale-a-video,venhancer} or edit videos \cite{edittempcons,fastyletrans}. Although these models are capable of producing realistic and exquisite videos, the lack of long video datasets \cite{webvid, internvid} and the enormous computational cost make it difficult for these models to generate long videos. Therefore, most of the existing methods are set to generate short videos no longer than $10$ seconds.

Due to a training-inference gap, directly applying existing short video diffusion models to long video generation will result in multiple forms of quality degradation. Diffusion models trained with short videos only learn short-term sequential dependencies and local spatio-temporal features during the training phase, with the receptive field and temporal modeling capabilities of its noise prediction network constrained within the length of the training videos. When generating video sequences beyond the training length during inference, the models face an out-of-distribution extrapolation problem, failing to effectively capture long-term temporal consistency and global motion patterns. \cite{freelong} found that directly applying short video diffusion models to long video generation leads to the distortion of high-frequency content in the temporal domain. This results in over-smoothing and temporal flickering issues. Some studies have proposed various training-free methods for long video generation, but they exclusively focus on either long-range temporal consistency \cite{freelong,freenoise} or motion diversity \cite{fifo}. These methods fail to acknowledge the necessity of evaluating the quality of long videos by considering both aspects simultaneously, and may produce long videos with restricted content or abrupt transitions. Besides, some methods expand the length of generated videos by modifying temporal attention layers, and are only suitable for video diffusion models that have separate temporal attention modules (2D+1D attention modules). For models with 3D-attention modules, performing these operations may simultaneously destroy the coherence in spatial dimensions and affect the generated results. 

In this article, we propose Diffusion Video Fusion (Diff-VF) for video diffusion models to tackle these challenges. Diff-VF contains three strategies, taking into account both temporal coherence and motion diversity. Firstly, we introduce hybrid noise initialization (HNI) to constrain global semantics while simultaneously ensuring rich contents. We use the first noise clip to determine the general content of the long video, and add additional noise in subsequent clips to enrich the content. Secondly, we introduce weighted window sampling (WWS) to remove inter-window discontinuities and maintain motion smoothness. We split a long video into clips with overlaps and give a different weight for each frame when merging the clips together. Thirdly, to establish long-range dependencies and ensure global consistency, we propose temporal extended sampling (TES). We divide the video into clips, each containing frames that have the same time interval in the original video. Then after the denoising process, we rearrange all frames in the original order. This mechanism maintains the consistency between distant frames during the denoising process, thereby ensuring long-range consistency. Furthermore, to apply this framework to video enhancement, we introduce the Skip Residual Guidance mechanism, which fuses the original video latents with denoised latents after each denoising step to maintain the fidelity of the enhanced video. The aforementioned strategies are all designed to function in latent space, which is independent of model architectures. The effectiveness of Diff-VF is illustrated in \cref{teaser}, where the proposed method generates long videos with $4 \times$ extended duration compared with the original videos produced by the base model.

Our contribution can be summarized as follows: \textbf{1)} We propose a training-free, plug-and-play and model-agnostic framework Diff-VF to use existing short video diffusion models to generate high-quality long videos. Our framework maintains long-range consistency and overall quality of the video while introducing variant motion. \textbf{2)} We propose Skip Residual Guidance mechanism for long video enhancement to improve the fidelity between the original video and the enhanced video. \textbf{3)} We applied our method to two video diffusion models: LaVie \cite{lavie} and HunyuanVideo \cite{hunyuanvideo}. Evaluations on VBench-Long \cite{vbench} show that Diff-VF achieves a favorable consistency-diversity trade-off compared with recent training-free baselines, including FreeNoise \cite{freenoise}, FreeLong \cite{freelong} and RIFLEx \cite{riflex}. In particular, Diff-VF preserves competitive temporal coherence and visual quality while reducing the tendency toward static or repetitive long-video generation. Moreover, the effectiveness of our proposed method has been demonstrated in both base models, incorporating 2D+1D and 3D attention modules.

\section{Related Works}
\label{related}

\subsection{Video Diffusion Models}

Diffusion models (DMs) employ two opposing processes to establish the relationship between images and noise \cite{sohl2015, ddpm, score}. The forward process adds noise into images step by step, while the reverse process uses the model to predict the noise added before and iteratively removes it to reconstruct clean images. The structure of DMs may be U-Net \cite{ddpm,ddim,ldm}, Transformer \cite{dit}, or other models. Latent diffusion models (LDMs) \cite{ldm} employ an autoencoder to transform images into latent features and implement the diffusion process in latent space. This reduces the memory and time cost in the training and inference stage, and enables the model to take multiple kinds of information as guidance.

To apply diffusion process in video generation, most methods choose to take the image generation model Stable Diffusion \cite{ldm} as the backbone, and integrate temporal modules into the layers of U-Net. In this way, they can fully leverage the diffusion prior of the pre-trained model, and compensate for the lack of video training data. Some other methods have explored ways to generate videos with higher quality \cite{sora, cogvideo, cogvideox, hunyuanvideo,wan,textdriven}. However, the lack of high-quality long video datasets and enormous memory cost of video latents in model training make it hard to train these models with sufficient long videos. Consequently, most video diffusion models are only able to generate short videos, and may suffer severe degradation when extending the length of the video. 

Recent works further investigate long video generation in more task-specific scenarios, especially for subject-consistent human video synthesis. OmniHuman-1.5 \cite{omnihuman15} introduces an audio- and image-conditioned avatar generation framework, where a pseudo last-frame condition with shifted RoPE is used to preserve the reference identity while avoiding excessive constraints on dynamic motion. InfiniteTalk \cite{infinitetalk} studies sparse-frame video dubbing and employs temporal context frames together with sparse reference keyframes to support infinite-length audio-driven generation with consistent identity, background, and camera motion. These methods demonstrate the importance of explicit identity and motion conditioning for long video synthesis, but they are mainly designed for human-centric or audio-driven scenarios with additional reference signals.

In addition, video diffusion models can be applied in video enhancement. Compared to previous methods \cite{realesrgan,realbasicvsr}, the generative power of diffusion models greatly benefits the overall quality of the enhanced videos. Some methods \cite{upscale-a-video,lavie,seine} focus on enhancing a single aspect of video quality, such as spatial resolution or framerate, while others \cite{i2vgen,venhancer} are able to enhance multiple dimensions of quality at one time. Similar to video generation, these models are initially trained to enhance short videos, and the same issues arise when these models are applied to long video enhancement.

\begin{figure}[t]
    \centering
    \includegraphics[width=\linewidth]{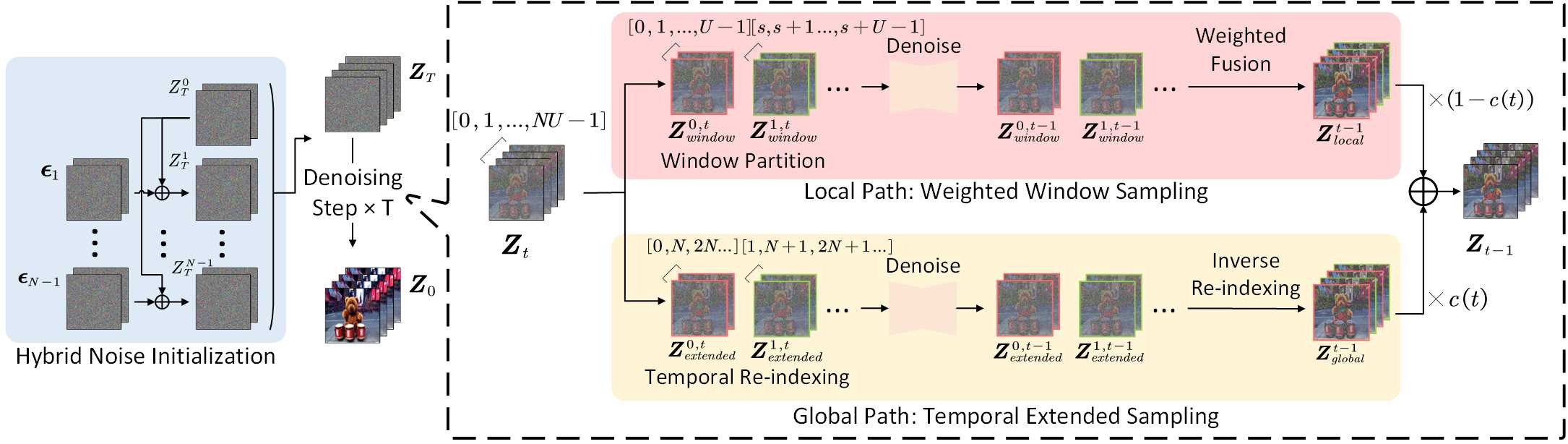}
    \caption{Overview of Diff-VF. We first build the initial noise through hybrid noise initialization, which replicates the first sequence of initial noise and mix with random noise to construct subsequent noise clips. Then in the denoising steps, we conduct weighted window sampling and temporal extended sampling and combine them. In weighted window sampling, the video is split into windows with overlaps and denoised separately. The windows are then aggregated together, where each frame in each window has its own weight. In temporal extended sampling, the latent sequence is temporally re-indexed into interleaved clips with a fixed temporal interval. After denoising, inverse temporal re-indexing restores the original chronological order. Finally, we use a video decoder with temporal layers to decode the latents into videos.}
    \label{overall}
\end{figure}

\subsection{Training-free Long Video Generation with Diffusion Models}

Researchers have applied different methods to fully leverage the generative power of existing video diffusion models in long video generation.  \cite{freeinit, freenoise, freelong, cono, confiner} emphasize the importance of initial noise in video consistency. They replicate or shuffle the noise to keep the long video consistent while applying short video diffusion models. 

To alleviate the degradation caused by temporal modules, \cite{gen-l-video} split the video into windows and process separately. \cite{freenoise,freelong} adopt similar algorithms, but they perform such operations in the temporal layers of diffusion models. This strategy is useful when video diffusion models have separate modules for spatial and temporal modeling \cite{lavie,videocrafter2}. However, for some other models \cite{i2vgen, hunyuanvideo, cogvideox}, the spatial and temporal modules are coupled together. Therefore, the imaging quality of the generated videos may be affected when such strategies are applied to these models, while our method solely operates on the noisy latent representations, remaining invariant to different spatial-temporal modeling strategies.

To improve the coherence between adjacent video clips,  \cite{cono,confiner} use L2 loss to amend the predicted noise of the next clip, making it more consistent with the previous clip.

FIFO-based methods generate long videos through a fixed-length latent queue, in which different queue elements are assigned shifted diffusion timesteps. At each iteration, the cleanest latent is dequeued from the head, while a newly initialized noisy latent is appended to the tail \cite{fifo}. However, vanilla FIFO-Diffusion may suffer from long-range inconsistency because it lacks explicit correspondence modeling across frames. Ouroboros-Diffusion \cite{ouroboros} further improves this paradigm through coherent tail latent sampling, Subject-Aware Cross-Frame Attention, and self-recurrent guidance, thereby enhancing structural and subject consistency.

\cite{riflex, longdiff} found that the frequency of temporal positional embeddings greatly affects the repetition patterns in generated videos. By specially designing positional embeddings in long video generation, the motion in long videos can be controlled in a reasonable range, and more acceptable for original short video diffusion models to manage temporal dependencies. However, for longer videos, such specially designed embeddings will prevent diffusion models from producing new content, limiting the diversity in generated long videos.

Unlike the above methods that overemphasize either temporal consistency or content diversity, our method considers both aspects and integrates multiple complementary strategies to achieve a better balance between them.

\section{Methodology}
\label{method}

Diff-VF is a general framework to generate high-quality long videos with short video diffusion models. As illustrated in \cref{overall}, we first perform hybrid noise initialization to get initial noise. In each denoising step, we use two strategies to do sampling and combine the results based on a weight that varies with timesteps. Finally, we use the VAE decoder in Stable Video Diffusion \cite{lvdm} to decode the denoised latent to further ensure the coherence of videos.

\subsection{Hybrid Noise Initialization}

DDIM \cite{ddim} introduced a non-Markovian diffusion process where the denoising process is entirely deterministic, dependent only on the initial noise and free from stochastic components. Building on this idea, FreeNoise \cite{freenoise} and FreeLong \cite{freelong} constructed the initial noise of the entire video by replicating the first video clip, introducing randomness by shuffling the frames in subsequent clips.

However, the shuffling process introduces only limited variability, and when applied to long video generation, it results in a repetition of the original short video, which is undesirable for generating diverse and coherent long videos. In response, we propose hybrid noise initialization, an innovative and effective approach that introduces additional random noise into the first noise clip's repetition. This strategy enhances the diversity of generated video content while ensuring global semantic consistency. Specifically, we first generate $\boldsymbol{Z}_T^0$ as the initial noise of the first video clip. After that, we mix the basic noise with newly generated random noise $\boldsymbol{\epsilon}_n \sim \mathcal{N}(\boldsymbol{0},\boldsymbol{I})$ to construct the subsequent clips, until we get all noise clips of the long video:
\begin{equation}
    \boldsymbol{Z}_T^n = \sqrt{1-w}\cdot \boldsymbol{Z}_T^0 + \sqrt{w}\cdot\boldsymbol{\epsilon}_n, \quad n=1,...,N-1,
\end{equation}
where $N$ is the number of short video clips in a long video. $w$ is the weight of additional noise in subsequent clips. While larger $w$ enriches the video content, it inversely affects long-range consistency. After that, we split the frames in subsequent clips into small groups containing $N$ frames and swap the frames in each group:

\begin{equation}
    \boldsymbol{Z}_T^{n'}[b+a\cdot N] = \boldsymbol{Z}_T^{n}[(b+n) \bmod N + a\cdot N]
\end{equation}

By replicating the first sequence of initial noise as the basis of a long video, the consistency of the video is maintained, while the introduction of additional random noise facilitates the generation of more diverse motion in the video. The parameter $w$ provides a flexible control mechanism, allowing the balance between global consistency and local diversity to be finely tuned. By adjusting $w$, the method offers the ability to achieve an optimal trade-off, where larger values of $w$ increase the diversity of motion at the cost of some loss in long-range consistency, while smaller values prioritize global coherence.

\subsection{Weighted Window Sampling}

In image generation, some studies \cite{multidiffusion, demofusion} have adopted sliding-window sampling approach. They split images into small crops with overlaps, perform denoising process one by one and fuse them together to reconstruct the original image. In this way they are able to generate coherent images with ultra-high resolution while maintaining an acceptable level of memory cost. 

\begin{figure}
    \centering
    \subfloat[simple fusion]{
        \includegraphics[width=0.45\linewidth]{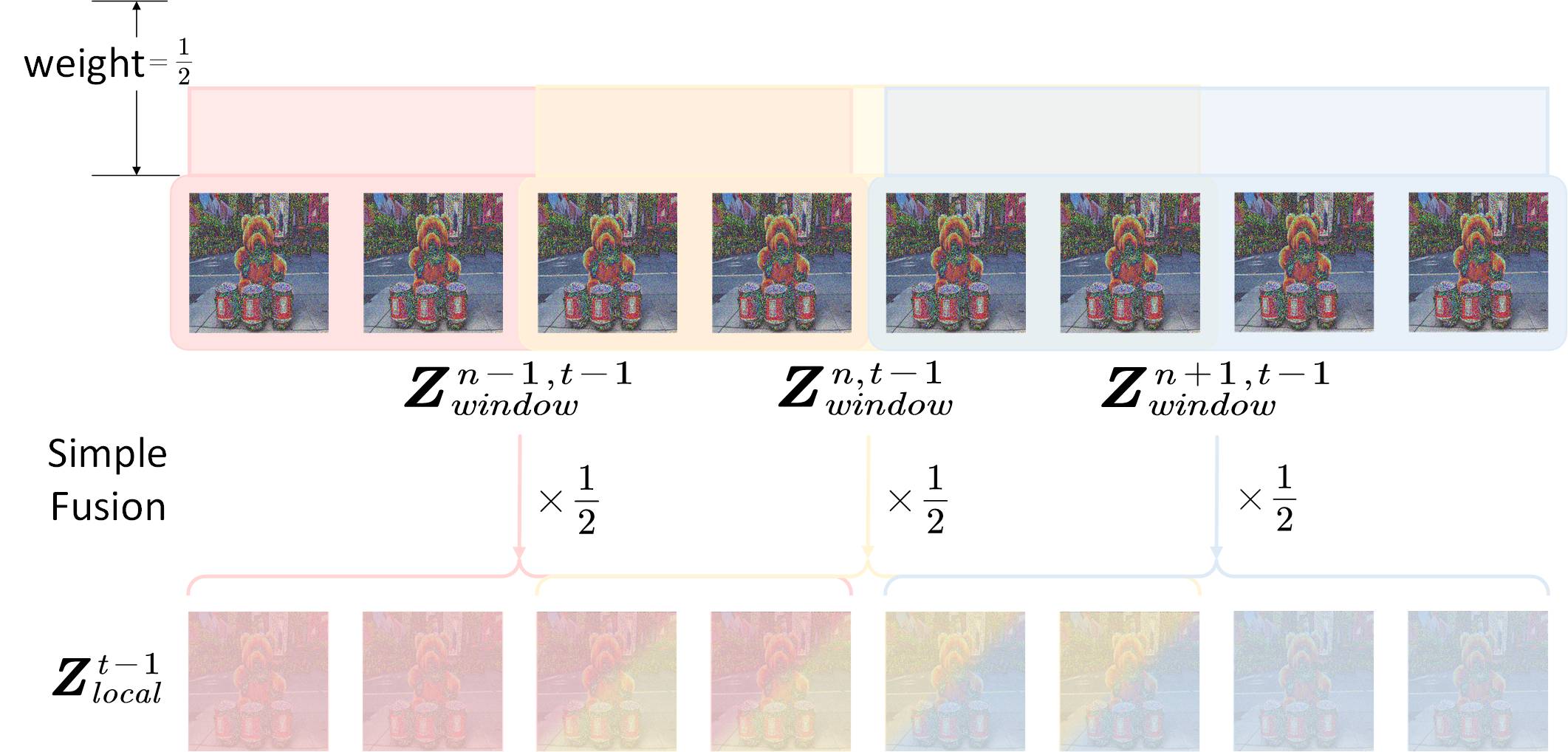}
        \label{sf}
    } 
    \subfloat[weighted fusion]{
        \includegraphics[width=0.45\linewidth]{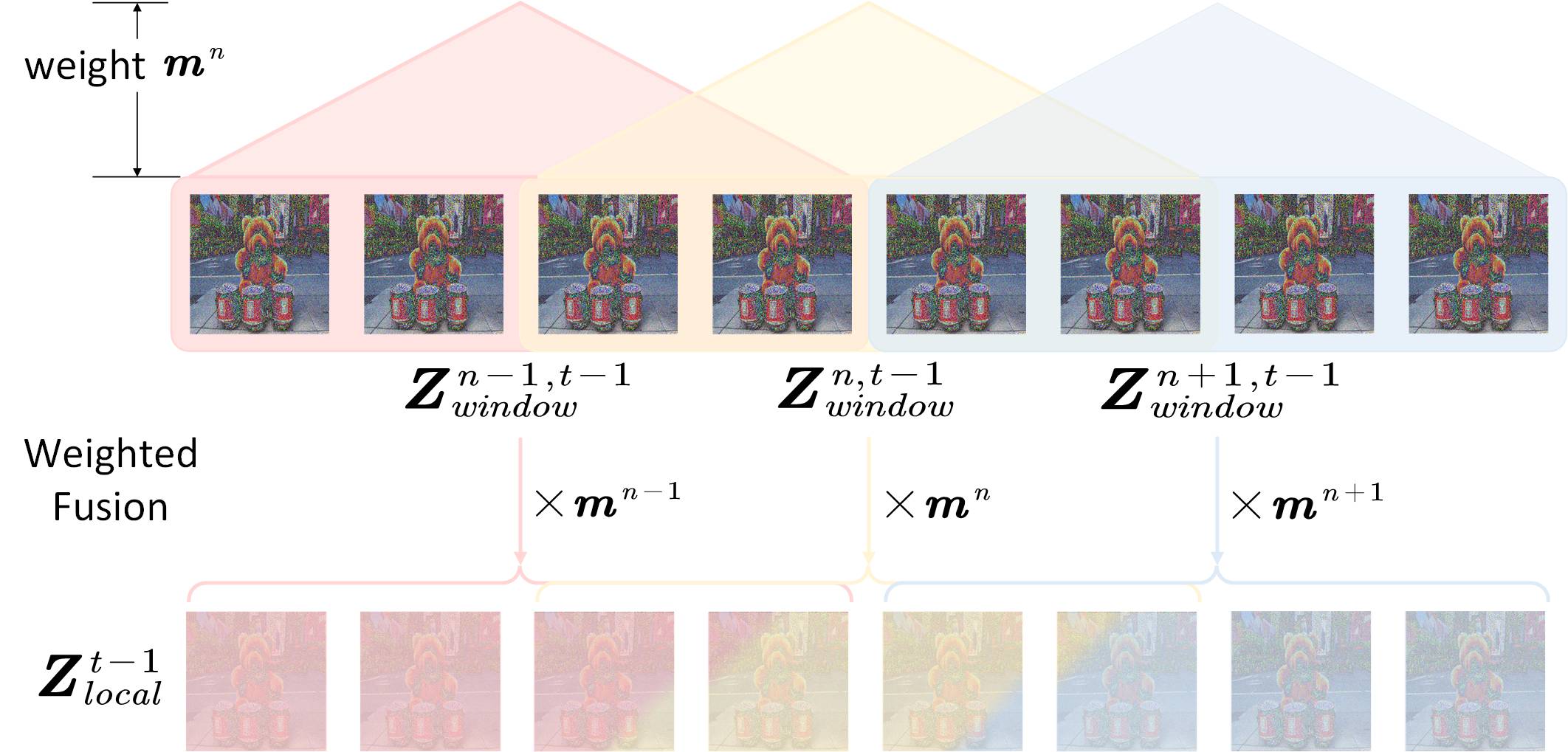}
        \label{wf}
    }
    \caption{Detailed illustration of simple fusion and weighted fusion. (a) If we simply calculate the average of the same frames in different windows, there will be significant differences between adjacent frames at the boundaries of windows, which leads to abrupt transitions. (b) We choose to set a varying weight to each frame in the windows. The weights of the frames in the previous window decreases gradually, while the weights in the next window increases. This strategy eliminates the distinct boundaries between windows, and achieves temporally smooth transition.}
    \label{sw}
\end{figure}

The enhancement of video length in the temporal dimension is analogous to the enhancement of image resolution in the spatial dimension. It is a natural way to split the video into windows and process them separately. We define these windows as:
\begin{equation}
    \boldsymbol{Z}_{window}^{n, t} = \boldsymbol{Z}_t[n \cdot s:n \cdot s+U], \;n=0,1,...
\end{equation}

where $s$ denotes the sampling stride between windows and $U$ denotes the window size. After all these windows are denoised to $\boldsymbol{Z}_{window}^{n,t-1}$, the simplest method to aggregate them back into a long video $\boldsymbol{Z}_{local}$ is to sum them together and calculate the average of different windows to get the frames in overlaps. Each frame $\boldsymbol{Z}_{local}^{t-1}[i]$ can be calculated through:

\begin{equation}
    \boldsymbol{Z}_{local}^{t-1}[i] = \frac{\sum_{j}^{}\boldsymbol{Z}_{window}^{j,t-1}[i]}{\sum_{j}1} \label{eq3}
\end{equation}
where $j$ denotes the indexes of the windows that contain the $i$-th frame. As shown in \cref{sf}, we call this simple fusion.

However, the differences between calculations of adjacent windows will lead to significant discontinuities (such as the 2nd and 3rd frames, 4th and 5th frames, 6th and 7th frames in \cref{sf}), and result in abrupt changes in the video. Therefore, we propose a novel weighted fusion strategy. We give an index-varying weight $m_{i,j} = \frac{U+1}{2}-\lvert C_j - i \rvert$ to the $i$-th frame in the $j$-th window, and $C_j$ is the central frame index in the $j$-th window.

\begin{equation}
    \boldsymbol{Z}_{local}^{t-1}[i] = \frac{\sum_{j}^{} m_{i,j}\boldsymbol{Z}_{window}^{j,t-1}[i]}{\sum_{j}^{} m_{i,j}} \\
\end{equation}

This weight function is designed to gradually smooth the transition between overlapping windows, ensuring that adjacent frames blend more naturally. Specifically, as illustrated in \cref{wf}, in the overlapping regions of two consecutive windows, the weight of the earlier window gradually decreases while the weight of the later window increases. This dynamic adjustment creates temporal continuity by alleviating abrupt changes and ensuring smooth transitions between frames.

\subsection{Temporal Extended Sampling}

Although our hybrid noise initialization strategy preliminarily keeps the similarity among all video clips, the additional random noise may still weaken the connection between different clips, and introduce undesired transitions after denoising.

The dilation of receptive field in neural networks has been proven to be an effective technique in high-resolution image generation \cite{scalecrafter, demofusion}. By increasing the receptive field, these models are able to integrate global context in their denoising processes, which enables them to capture relationships between distant pixels. Building on this concept, we adapt receptive field dilation to the temporal dimension and propose temporal extended sampling as a temporal re-indexing strategy for long video generation.

It is worth noting that temporal extended sampling is not a two-stage key-frame generation strategy. It does not first synthesize sparse key frames, nor does it use them as hard conditional anchors for the remaining frames. Instead, at each denoising step, it reorganizes the current noisy latent sequence into several temporally dilated clips, applies the original short-video diffusion backbone to these clips, and then scatters the denoised latents back to their original temporal positions.

Specifically, given the long-video latent $\boldsymbol{Z}_t$, we divide its temporal indices into $N$ interleaved subsequences. Each subsequence contains $U$ frames with a fixed temporal interval $N$ in the original long video. For the $n$-th subsequence, where $n=0,1,\ldots,N-1$, the temporal extended clip is constructed as
\begin{equation}
    \boldsymbol{Z}_{extended}^{n, t}
    =
    \boldsymbol{Z}_t[n+k\cdot N],
    \; k=0,1,\ldots,U-1 .
\end{equation}
In this way, adjacent frames in $\boldsymbol{Z}_{extended}^{n,t}$ are not adjacent in the original video, but are sampled from temporally distant positions. Therefore, when the frozen short-video diffusion model denoises such a reorganized clip, its original temporal modeling ability is reused to establish dependencies among distant segments of the long video.

After all temporal extended clips are denoised, we scatter the frames back according to the inverse temporal re-indexing operation:
\begin{equation}
    \boldsymbol{Z}_{global}^{t-1}[n+k\cdot N]
    =
    \boldsymbol{Z}_{extended}^{n, t-1}[k],
    \quad
    n=0,1,\ldots,N-1,\;
    k=0,1,\ldots,U-1 .
\end{equation}
This produces a global-update path $\boldsymbol{Z}_{global}^{t-1}$, in which each frame has been updated through a temporally dilated clip. Temporal extended sampling thus extends the effective receptive field of short video diffusion models to include frames from different segments of a long video. By incorporating this temporal dilation, the denoising process can better capture long-range temporal dependencies and improve global consistency across the entire video.

After weighted window sampling and temporal extended sampling are performed, the results from the two paths are fused with a timestep-varying weight:
\begin{equation}
    \boldsymbol{Z}_{t-1}
    =
    (1-c(t))\cdot \boldsymbol{Z}_{local}^{t-1}
    +
    c(t) \cdot \boldsymbol{Z}_{global}^{t-1}.
\end{equation}
where
\begin{equation}
    c(t)
    =
    \alpha \cdot
    \left(
    \frac{1}{2}
    \left(
    1+\cos\left(\pi \cdot \frac{T-t}{T}\right)
    \right)
    \right)^{c_s},
    \quad
    \alpha \in [0,1].
\end{equation}
The global path produced by temporal extended sampling is therefore not used as a hard constraint, but is softly fused with the local path produced by weighted window sampling. This timestep-dependent design allows temporal extended sampling to play a more important role in establishing global structures in the early stage of denoising, while gradually reducing its influence in later steps, where the model mainly focuses on local motion continuity and textural details.

\subsection{Skip Residual Guidance}

To apply short video diffusion models to long video enhancement, we propose skip residual guidance, which ensures fidelity by integrating the information of original videos in the denoising process. We first use a lightweight net \cite{realesrgan} for super-resolution to primarily enhance the resolution of videos. Then the videos are encoded into latent space as $\boldsymbol{Z}_{input}$, and fused with $\boldsymbol{Z}_{T}$ generated by hybrid noise initialization. The noised latent is set to the state of timestep $T_{max}$. In the denoising process, we add noise into $\boldsymbol{Z}_{input}$ to align its noise level with $\boldsymbol{Z}_{t-1}$. After that we combine the noised input latent $\boldsymbol{Z}_{input}^{t-1}$ with $\boldsymbol{Z}_{t-1}$. This mechanism is executed in every step, and controlled by a weight which is relevant to the timestep: 

\begin{equation}
    \tilde{\boldsymbol{Z}}_{t-1} = (1-c'(t)) \cdot \boldsymbol{Z}_{t-1} + c'(t) \cdot \boldsymbol{Z}_{input}^{t-1}
\end{equation}

where $c'(t) =(\frac{1}{2}(1+\cos(\pi \cdot \frac{T-t}{T}))^{c_r}$. The skip residual guidance introduces an additional pathway to guide the generation of the enhanced video with the original video. We set a larger $c'(t)$ in early denoising steps to ensure the fidelity of enhanced videos, and gradually reduce it in the denoising process to allow diffusion models to add more realistic details. Thus we can achieve a balance between fidelity and quality when enhancing long videos with short video diffusion models by simply modifying the procedure of denoising steps, without the need of extra tunable modules.

\subsection{Differences from Existing Training-free Methods}
\label{sec:difference_existing_methods}

Although recent training-free methods have explored different ways to extend short-video diffusion models to long video generation, Diff-VF differs from them in the operating space and temporal organization of the denoising process. Specifically, HNI, WWS, and TES are all performed on noisy latent sequences, without modifying the parameters or internal modules of the pretrained backbone.

\paragraph{Hybrid Noise Initialization.}
Existing noise-based methods, such as FreeNoise~\cite{freenoise}, mainly construct long-video initial noise by reusing, shuffling, or rescheduling noise sequences. In contrast, HNI formulates the initial noise of each subsequent clip as a mixture of the first clip noise and newly sampled Gaussian noise. Therefore, the relationship among video clips is controlled by the explicit mixing weight $w$, rather than solely determined by replicated or rescheduled noise patterns.

\paragraph{Weighted Window Sampling.}
Window-based operations have been used in prior training-free long-video methods, such as window-based temporal attention or global-local temporal processing~\cite{freenoise, freelong}. The main distinction of WWS is its position-dependent aggregation rule for overlapping windows. After each local window is denoised, WWS assigns larger weights to frames near the window center and smaller weights to frames near the window boundary. Thus, predictions from adjacent windows are fused with gradually varying contributions in the overlapping region, rather than by uniform averaging. This weighted fusion is applied to latent window outputs and does not require modifying the temporal attention structure of the backbone.

\paragraph{Temporal Extended Sampling.}
Some methods extend video length by modifying positional encodings, such as RIFLEx~\cite{riflex}, or by maintaining a first-in-first-out denoising queue, such as Ouroboros-Diffusion~\cite{ouroboros}. TES adopts a different formulation. It neither changes the positional encoding nor performs queue-based generation. Instead, at each denoising step, TES temporally re-indexes the current long-video latent into several interleaved subsequences, forms temporally dilated clips, denoises them with the frozen backbone, and scatters the updated latents back to their original temporal positions.

\paragraph{Latent-space local-global fusion.}
The three components are integrated as a latent-space local-global denoising framework. HNI defines the initial relationship among different clips, WWS provides a local denoising path over adjacent temporal windows, and TES provides a global denoising path over temporally dilated clips. The local and global paths are then softly fused with the timestep-dependent weight $c(t)$. Since all operations are performed on latent sequences outside the model architecture, Diff-VF can be applied to different video diffusion backbones without requiring architecture-specific modifications.

\section{Experiments}
\label{experiments}

\begin{table}[t]
  \centering
  \caption{Quantitative comparison results of long video generation methods on LaVie. \textbf{Bold} and \underline{underline} stand for the best and second best results respectively.}
  \label{quantitative}
  \begin{tabular*}{\textwidth}{@{\extracolsep{\fill}} l cccccccc @{}}
    \toprule
    Method & \makecell{Subject\\Cons.} & \makecell{Motion\\Smooth.} & \makecell{Temp.\\Flick.} & \makecell{Dyn.\\Deg.} & \makecell{Overall\\Cons.} & \makecell{Human\\Action} & \makecell{Imag.\\Qual.} & \makecell{Aesth.\\Qual.} \\
    \midrule
    Base Model \cite{lavie} & 0.8546 & 0.9107 & 0.9289 & 0.9722 & 0.2609 & 0.9340 & 0.6076 & 0.5231 \\
    \midrule
    FreeNoise \cite{freenoise} & 0.9106 & 0.9485 & \underline{0.9856} & \underline{0.6958} & \underline{0.2745} & \underline{0.9340} & \underline{0.6405} & 0.5638 \\
    FreeLong \cite{freelong} & \textbf{0.9285} & \underline{0.9502} & \textbf{0.9899} & 0.5625 & 0.2697 & 0.9120 & \textbf{0.6467} & \textbf{0.5735} \\
    Ours & \underline{0.9164} & \textbf{0.9529} & 0.9848 & \textbf{0.7208} & \textbf{0.2795} & \textbf{0.9580} & 0.6368 & \underline{0.5712} \\
    \bottomrule
  \end{tabular*}
\end{table}

\begin{table}[t]
  \centering
  \caption{Quantitative comparison results of long video generation methods on HunyuanVideo.}
  \label{quantitative-hunyuan}
  \begin{tabular*}{\textwidth}{@{\extracolsep{\fill}} l cccccc @{}}
    \toprule
    Method & \makecell{Subject\\Consistency} & \makecell{Motion\\Smoothness} & \makecell{Temporal\\Flickering} & \makecell{Dynamic\\Degree} & \makecell{Imaging\\Quality} & \makecell{Aesthetic\\Quality} \\
    \midrule
    Base model \cite{hunyuanvideo} & 0.9800 & 0.9956 & 0.9945 & 0.2813 & 0.5992 & 0.5798 \\
    \midrule
    RIFLEx \cite{riflex}           & \underline{0.9858} & \underline{0.9957} & \underline{0.9961} & 0.1413 & 0.5814 & 0.5584 \\
    FreeNoise \cite{freenoise}     & 0.9704 & 0.9931 & 0.9888 & \underline{0.4031} & \textbf{0.6646} & \textbf{0.6014} \\
    FreeLong \cite{freelong}       & \textbf{0.9913} & \textbf{0.9966} & \textbf{0.9973} & 0.0569 & 0.5441 & 0.4828 \\
    Ours                           & 0.9594 & 0.9920 & 0.9864 & \textbf{0.4413} & \underline{0.6292} & \underline{0.5987} \\
    \bottomrule
  \end{tabular*}
\end{table}

\subsection{Implementation Details}

\paragraph{Setting Up.} 

We conduct our experiment on a U-Net-based model (LaVie\cite{lavie}) and a DiT-based model (HunyuanVideo\cite{hunyuanvideo}) to demonstrate the generality of our method. Each method on the same base model is tested with the same sampling strategy. For LaVie, we set $U=16$, $s=8$ and $N=4$ to generate $4\times$ length videos, and set $w=0.1$, $\alpha=0.4$ and $c_s=6$ for better video quality. For HunyuanVideo, we set $U=32$, $s=16$ and $N=3$ in latent space to generate $3\times$ length videos, and set $w=0.2$, $\alpha=0.5$ and $c_s=6$. We perform temporal extended sampling for $t<900$ on HunyuanVideo to avoid significant semantic discrepancies between temporal extended clips, which could be caused by the excessively high $c(t)$ during early denoising steps, leading to noticeable incoherence in the generated videos.

For long video enhancement, we choose ms-vid2vid-xl \cite{i2vgen} as our baseline model. To enhance the videos generated by LaVie with our method, we set $U=32$, $s=16$ and $N=2$. The denoising process begins at $t=900$, and we set $w=0.3$, $\alpha=1$ and $c_s=5$.

\paragraph{Evaluation Metrics.} 

We evaluate the quality of generated videos in dimensions of temporal coherence, motion diversity and semantic consistency. To evaluate temporal coherence, we choose subject consistency, motion smoothness and temporal flickering in VBench\cite{vbench} to measure the temporal consistency and smoothness of long videos. To evaluate motion diversity, we choose dynamic degree in VBench. To evaluate semantic consistency, we choose overall consistency and human action to reflect how the semantics and motion of humans in videos match the text prompts. To evaluate frame-wise quality, we choose imaging quality and aesthetic quality to measure pixel-wise distortion and beauty value of the videos.

We use VBench-Long method to evaluate the generated long videos. VBench-Long split videos into clips and apply slow-fast approach to evaluate long-range temporal consistency.

\subsection{Quantitative Comparison}

We compare our method with other training-free long video generation with diffusion models: 1) directly sampling long videos with original short video diffusion models; 2) FreeNoise \cite{freenoise}, which introduced noise replication and shuffling to maintain video consistency and apply sliding-window algorithm in temporal attention layers; 3) FreeLong \cite{freelong}, which followed the noise initialization method in FreeNoise and introduced local-global attention decoupling to maintain both long-range consistency and visual quality; 4) RIFLEx \cite{riflex} , which reduces the intrinsic frequency of RoPE \cite{roformer} components to extend the length of videos generated by DiT-based models.

\begin{figure}[t]
    \centering
    \includegraphics[width=\linewidth]{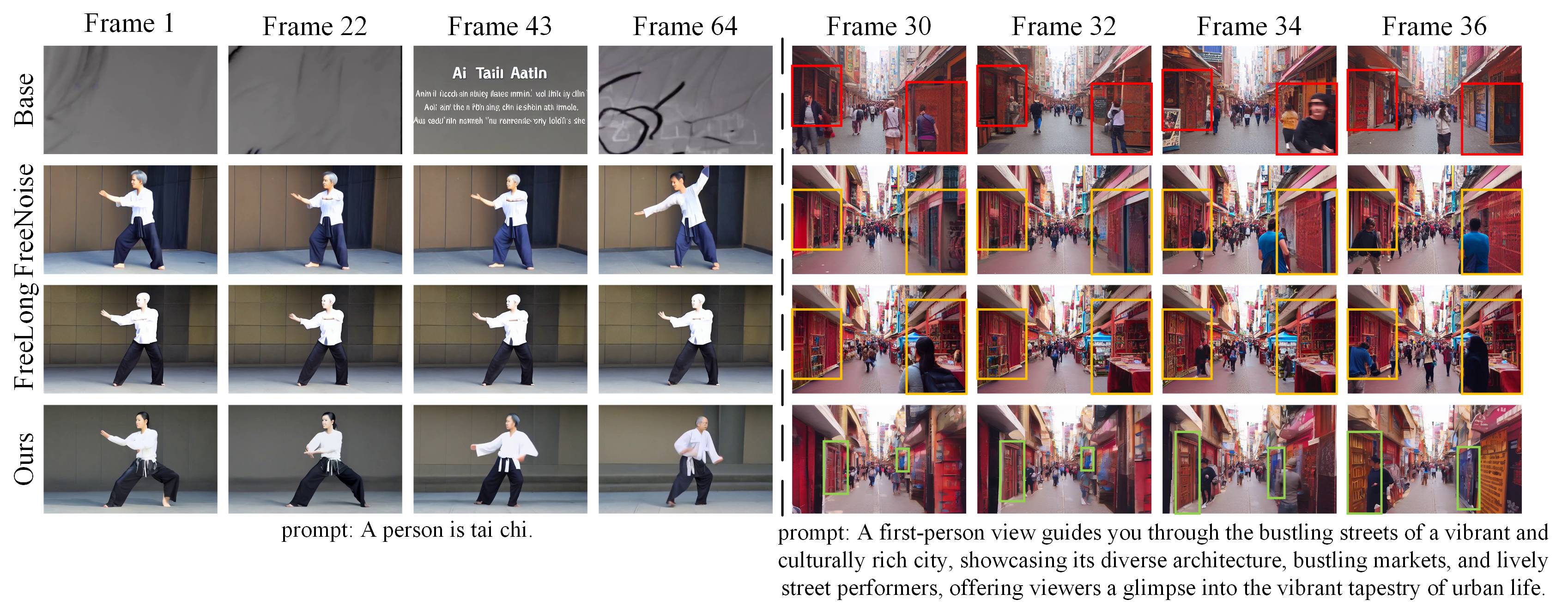}
    \caption{Qualitative comparison based on LaVie. Videos generated by the base model LaVie \cite{lavie} have over-smoothing and temporal flickering issues. As shown in the red boxes, even nearby frames have significant differences. Videos generated by FreeNoise \cite{freenoise} and FreeLong \cite{freelong} exhibit very limited dynamic range. The orange boxes show that the backgrounds in these videos are nearly unchanged. Videos generated by our method achieve a balance between temporal consistency and motion diversity, and perform better in semantic alignment. The backgrounds have significant movement over time, and the objects in the green boxes remain coherent in their movement. Complete videos corresponding to the visual results in this figure are provided in the supplementary material.}
    \label{qualitative}
\end{figure}

\cref{quantitative} presents the quantitative comparison results on LaVie. The videos used in evaluation are generated according to the standard prompt sets \cite{vbench}. 
The results in \cref{quantitative} indicate that long-video generation should be evaluated as a consistency-diversity trade-off rather than by individual metrics alone. The base model obtains a high dynamic degree, but this is accompanied by poor temporal flickering and visual instability, suggesting that the measured motion is partly caused by incoherent frame variations rather than meaningful object or camera motion. FreeNoise and FreeLong improve temporal consistency by increasing correlations among temporal segments, but their dynamic degree is significantly reduced, indicating that strong temporal correlation may suppress content evolution. In contrast, Diff-VF maintains competitive temporal consistency while preserving a higher level of motion dynamics, showing a more balanced behavior between long-range coherence and temporal variation. In addition, Diff-VF exceeds other methods in terms of semantic alignment and motion smoothness, demonstrating its ability to generate long videos with more coherent motions and closer adherence to text prompts.

\cref{quantitative-hunyuan} presents the quantitative comparison results on HunyuanVideo. The videos used in evaluation are generated according to a long-prompt set \cite{vbench} containing 40 prompts with detailed description. We used AccVideo \cite{accvideo} to accelerate the generation process. Due to the limits of VBench in evaluating videos generated with self-specified text prompts, we only choose six metrics.
Similar to the results on LaVie, methods such as RIFLEx and FreeLong achieve high subject consistency and temporal smoothness, but their dynamic degree is much lower. This suggests that part of the temporal consistency comes from periodic or nearly static generation, rather than from coherent long-range motion. FreeNoise produces stronger dynamics, but still tends to repeat contents due to its correlated noise design. Diff-VF obtains the highest dynamic degree among the compared methods while maintaining reasonable temporal and visual quality, which supports our goal of improving long-video generation from the perspective of consistency-diversity balance.

\subsection{Qualitative Comparison}

The qualitative results further reveal different failure modes of existing training-free long-video generation methods. Directly applying the base model to longer sequences often leads to unstable high-frequency details and temporal flickering, because the model is forced to denoise sequences beyond its training length. Noise-design-based methods, such as FreeNoise and FreeLong, commonly rely on strongly correlated initial noises to preserve the global appearance of long videos. By reusing, shuffling, or rescheduling noise sequences across temporal segments, these methods can improve long-range visual consistency, but the strong correlation may also bias different clips toward similar latent trajectories, leading to periodic content, repeated motions, or insufficient temporal evolution. In contrast, Diff-VF adopts Hybrid Noise Initialization to explicitly control the correlation between clips. Instead of only reusing the first noise sequence, HNI mixes it with newly sampled Gaussian noise, introducing clip-level stochasticity while preserving shared global semantics. This noise design provides a controllable balance between temporal consistency and motion diversity, and the subsequent WWS and TES paths further refine local smoothness and long-range dependencies in latent space.

\begin{figure}[t]
    \centering
    \includegraphics[width=0.9\linewidth]{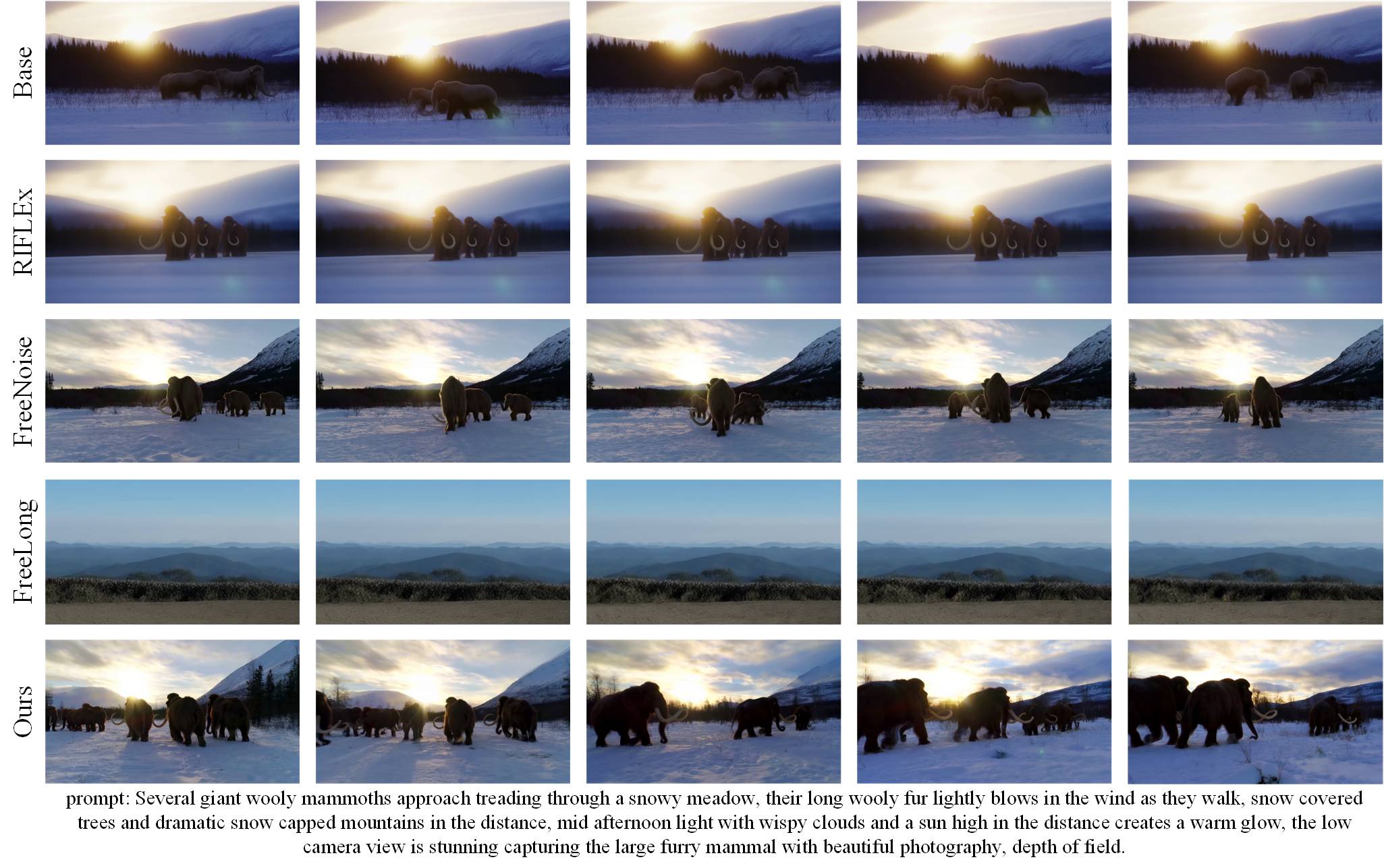}
    \caption{Qualitative comparison based on HunyuanVideo. Complete videos corresponding to the visual results in this figure are provided in the supplementary material.}
    \label{hunyuan_visual}
\end{figure}

\begin{figure}[t]
    \centering
    \includegraphics[width=0.9\linewidth]{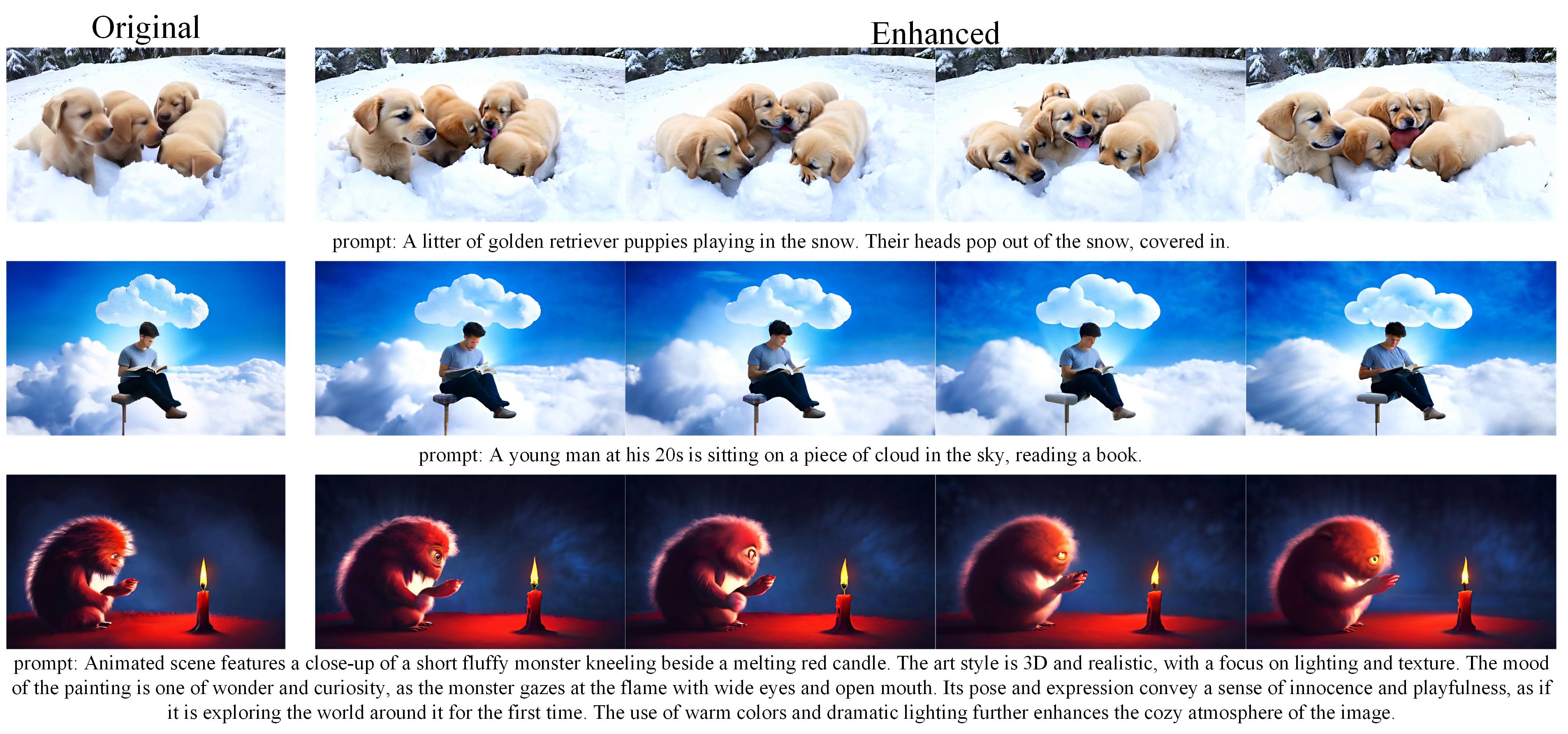}
    \caption{Qualitative results of our long video enhancement method. Compared to original videos, the enhanced videos have faithful and realistic details, and keep long-range consistency and fidelity at the same time. Complete input and enhanced video results corresponding to this figure are provided in the supplementary material.}
    \label{enhance_visual}
\end{figure}

In \cref{qualitative}, the base model shows noticeable frame-level instability. FreeNoise and FreeLong tend to produce visually consistent but weakly changing contents. Diff-VF generates more visible temporal evolution while preserving the identity and background structure of the scene. This observation is consistent with the quantitative trade-off in \cref{quantitative}. More results will be shown in the appendix.

In \cref{hunyuan_visual}, the difference between methods is more related to architectural compatibility. RIFLEx directly modifies positional extrapolation and can preserve temporal structure, but may still lead to repeated or slowed-down motion. FreeLong relies on temporal attention operations, which are less straightforward for DiT backbones with coupled spatial-temporal attention. Diff-VF avoids modifying internal attention modules and instead reorganizes latent frames outside the backbone, which helps preserve both visual fidelity and temporal evolution. More results will be shown in the appendix.

\cref{enhance_visual} shows some videos enhanced with our method and the corresponding original videos. It is evident that our method for long video enhancement incorporates faithful and realistic details into the videos while keeping long-range temporal consistency and fidelity at the same time.

Since still frames cannot fully reflect temporal continuity and motion dynamics, we provide the complete videos corresponding to all qualitative comparisons in the supplementary material.

\subsection{Ablation Studies}

\subsubsection{Ablations on Strategies}

In this part, we evaluate the effectiveness of each strategy used in our method. The ablation results provide further evidence for the functional decomposition of Diff-VF.

\paragraph{Ablation on Hybrid Noise Initialization (HNI)} 

\begin{figure}[t]
    \centering
    \includegraphics[width=\linewidth]{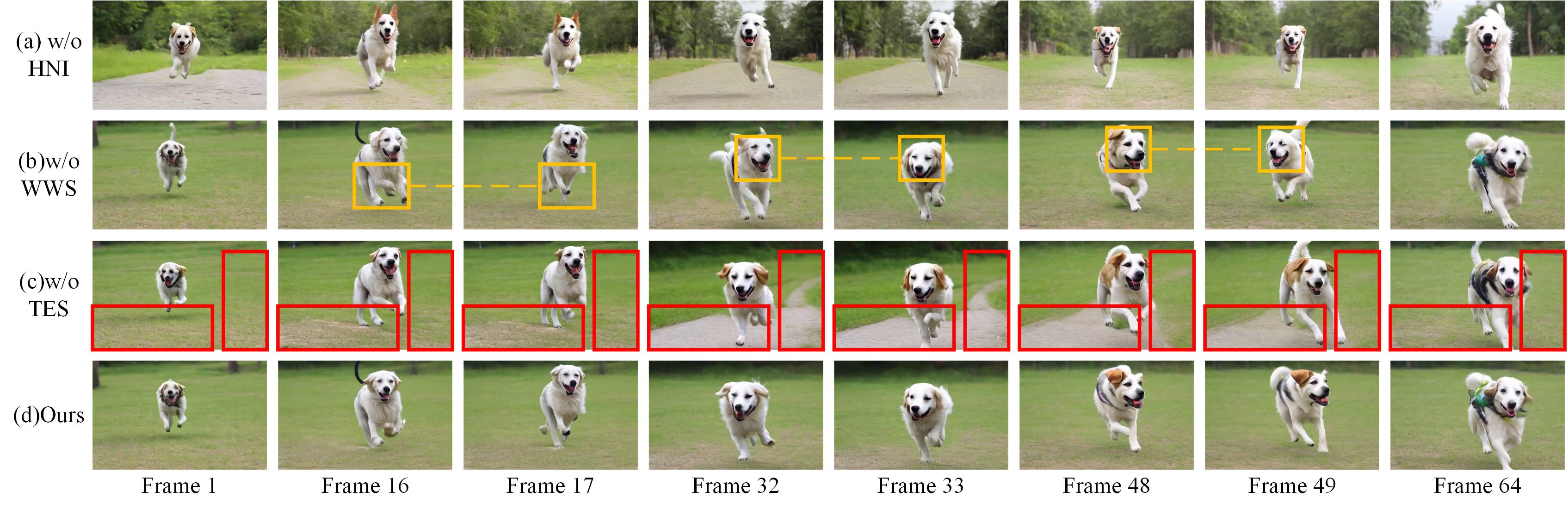}
    \caption{Visual comparison of ablation studies. Prompt: A dog running happily. (a) A video generated without hybrid noise initialization. Without the control of initial noise, the content varies significantly in different parts of the video. (b) A video generated without weighted window sampling. At the boundaries of video clips, the adjacent frames have apparent differences, which affects the smoothness of the whole video. (c) A video generated without temporal extended sampling. This video has smooth motion, but the background changes greatly as the video progresses. (d) A video generated by our method, which performs well in both temporal consistency and motion smoothness. Complete videos corresponding to the visual results in this figure are provided in the supplementary material.}
    \label{ablation_visual}
\end{figure}

Removing HNI replaces the structured clip-level noise relation with independent Gaussian noise, which weakens the semantic connection among different temporal segments. As a result, the generated video contains richer variations but suffers from reduced long-range consistency. As shown in \cref{ablation} and \cref{ablation_visual}, without hybrid noise initialization, our method performs poorly in long-range consistency, motion smoothness and temporal flickering. Thus we demonstrate that hybrid noise initialization ensures video consistency and smoothness by limiting video content in a reasonable range.

\paragraph{Ablation on Weighted Window Sampling (WWS)} 

Removing WWS keeps the global temporal path but replaces the position-dependent aggregation of overlapping local windows with simple aggregation, leading to abrupt transitions around window boundaries. As shown in \cref{ablation} and \cref{ablation_visual}, the videos generated without weighted window sampling have abrupt transitions at the boundaries of the windows, which affects the temporal consistency and motion smoothness of the video.

\paragraph{Ablation on Temporal Extended Sampling (TES)} 

Removing TES preserves local smoothness but weakens the interaction among distant temporal segments, so the video may remain locally continuous while drifting globally. As shown in \cref{ablation_visual}, without temporal extended sampling, the content changes greatly in different parts of the video. 

\medskip

Moreover, compared to other methods, our method reasonably blends different video clips together, thus avoiding abrupt transitions that affect frame-wise quality. Therefore our method performs the best in imaging quality and aesthetic quality. These observations suggest that HNI, WWS, and TES address complementary aspects of long-video generation: initial semantic correlation, local boundary smoothness, and long-range temporal dependency.

\subsubsection{Ablations on Hyperparameters}

In this part, we evaluate the influence of hyperparameters in our method, and explain how we choose them. 

\paragraph{Ablation on $\alpha$.} 

$\alpha$ controls the influence of weighted window sampling and temporal extended sampling. These two approaches respectively focus on maintaining local and global consistency. A larger $\alpha$ assigns a higher weight to temporal extended sampling, which is beneficial for enforcing long-range consistency; however, an excessively large $\alpha$ may introduce noticeable discontinuities between adjacent frames, thereby impairing motion continuity. Conversely, a smaller $\alpha$ improves short-range consistency and motion smoothness, but may cause the video content to drift substantially over time. As shown in \cref{ablation_hyper}, only by striking an appropriate balance ($\alpha=0.4$) can the overall coherence of the video be enhanced, and the best human action and imaging quality scores are achieved under this circumstance.

\paragraph{Ablation on $c_s$.} 

$c_s$ controls the degree of variation in the weights of weighted window sampling and temporal extended sampling over timesteps. Prior works \cite{ddpm,earlystage} have shown that diffusion models at earlier denoising steps are primarily responsible for establishing the global semantics of a video, while later steps mainly refine local details. Consequently, we choose a suitable $c_s$
to ensure that temporal extended sampling has a stronger influence at early denoising steps, thus promoting global temporal consistency, whereas weighted window sampling becomes more dominant at later denoising steps, leading to smoother visual details and more coherent motion. Similar to $\alpha$, \cref{ablation_hyper} shows that $c_s$ should be set to a reasonable value. An excessively large $c_s$ may undermine long-range consistency, whereas an overly small $c_s$ can impair motion smoothness. Taking into account the temporal consistency, dynamics and frame-wise quality of the video, we choose $c_s=6$ in our method.

\paragraph{Ablation on $w$.} 

$w$ controls the extent to which additional content is added in hybrid noise initialization. As shown in \cref{ablation_hyper}, A larger $w$ introduces greater variation in the content of each video clip; however, such variation may in some cases decisively alter the global semantics of individual clips, resulting in inconsistencies across the overall video. In contrast, a smaller $w$ favors temporal consistency, but the resulting differences among the initial noises of video segments may be insufficient to induce perceptible changes in the generated content, causing long videos to degenerate into monotonous repetitions of similar clips. Taking into account the temporal consistency, dynamics and frame-wise quality of the video, we choose $w=0.1$ in our method.

The ablation of HNI and $w$ can also be interpreted as an analysis of noise design. Independent Gaussian noise introduces strong clip-level stochasticity and thus increases content variation, but it weakens the semantic correlation among clips and leads to worse long-range consistency. In contrast, fully replicated or strongly correlated noise, as used by noise-rescheduling baselines, stabilizes appearance but may cause repetitive content. HNI lies between these two extremes by explicitly controlling the correlation between the first noise clip and subsequent clips. The trend of $w$ further verifies this interpretation: increasing $w$ injects more stochastic content and improves motion diversity, but excessive stochasticity decreases subject consistency and frame-wise quality. Therefore, $w$ serves as a direct control knob for the noise-induced consistency-diversity trade-off.

\medskip

The hyperparameter ablations further show that Diff-VF is controlled by explicit trade-off parameters. The noise mixing weight $w$ controls the strength of clip-level stochasticity in HNI. A small $w$ makes different clips highly correlated, while a large $w$ increases clip-level variation. The fusion weight scale $\alpha$ and schedule exponent $c_s$ control the relative contribution of the global TES path and the local WWS path across denoising steps. Therefore, these hyperparameters do not independently optimize a single metric; instead, they determine the balance among temporal consistency, motion diversity, and local visual smoothness.

\begin{table}[t]
  \centering
  \caption{Quantitative results of ablation studies on strategies}
  \label{ablation}
  \begin{tabular*}{\textwidth}{@{\extracolsep{\fill}} l cccccccc @{}}
    \toprule
    Method & \makecell{Subject\\Cons.} & \makecell{Motion\\Smooth.} & \makecell{Temp.\\Flick.} & \makecell{Dyn.\\Deg.} & \makecell{Overall\\Cons.} & \makecell{Human\\Action} & \makecell{Imag.\\Qual.} & \makecell{Aesth.\\Qual.} \\
    \midrule
    Ours w/o HNI & 0.8868 & 0.9429 & 0.9733 & \textbf{0.8368} & \textbf{0.2880} & \textbf{0.9600} & 0.6308 & 0.5629 \\
    Ours w/o WWS & \underline{0.9123} & 0.9478 & \textbf{0.9867} & 0.7367 & 0.2789 & 0.9460 & 0.6344 & \underline{0.5704} \\
    Ours w/o TES & 0.9118 & \textbf{0.9551} & 0.9824 & \underline{0.7590} & \underline{0.2810} & \textbf{0.9600} & \underline{0.6350} & 0.5691 \\
    Ours         & \textbf{0.9164} & \underline{0.9529} & \underline{0.9848} & 0.7208 & 0.2795 & \underline{0.9580} & \textbf{0.6368} & \textbf{0.5712} \\
    \bottomrule
  \end{tabular*}
\end{table}

\begin{table}[t]
  \centering
  \caption{Ablation studies on hyperparameters}
  \label{ablation_hyper}
  \begin{tabular*}{\textwidth}{@{\extracolsep{\fill}} cc cccccccc @{}}
    \toprule
    \multicolumn{2}{c}{Hyperparameters} & \makecell{Subject\\Cons.} & \makecell{Motion\\Smooth.} & \makecell{Temp.\\Flick.} & \makecell{Dyn.\\Deg.} & \makecell{Overall\\Cons.} & \makecell{Human\\Action} & \makecell{Imag.\\Qual.} & \makecell{Aesth.\\Qual.} \\
    \midrule
    \multirow{4}{*}{$\alpha$} 
      & 0.2        & 0.9147 & \textbf{0.9545} & 0.9837 & \textbf{0.7423} & \textbf{0.2805} & 0.9520 & 0.6358 & 0.5709 \\
      & 0.4 (Ours) & \textbf{0.9164} & \underline{0.9529} & 0.9848 & \underline{0.7208} & \underline{0.2795} & \textbf{0.9580} & \textbf{0.6368} & 0.5712 \\
      & 0.6        & \underline{0.9159} & 0.9504 & \underline{0.9857} & 0.7090 & 0.2790 & \underline{0.9540} & \underline{0.6363} & \underline{0.5723} \\
      & 0.8        & 0.9138 & 0.9468 & \textbf{0.9863} & 0.7118 & 0.2786 & 0.9500 & 0.6360 & \textbf{0.5725} \\
    \midrule
    \multirow{4}{*}{$c_s$} 
      & 2          & 0.9161 & 0.9495 & \textbf{0.9865} & 0.7160 & 0.2789 & 0.9520 & 0.6342 & \underline{0.5716} \\
      & 4          & \textbf{0.9166} & 0.9518 & \underline{0.9854} & 0.7139 & 0.2794 & \underline{0.9540} & 0.6359 & \textbf{0.5722} \\
      & 6 (Ours)   & \underline{0.9164} & \underline{0.9529} & 0.9848 & \underline{0.7208} & \underline{0.2795} & \textbf{0.9580} & \textbf{0.6368} & 0.5712 \\
      & 10         & 0.9157 & \textbf{0.9539} & 0.9841 & \textbf{0.7354} & \textbf{0.2803} & 0.9520 & \underline{0.6364} & 0.5710 \\
    \midrule
    \multirow{4}{*}{$w$} 
      & 0.05       & \textbf{0.9187} & \textbf{0.9535} & \textbf{0.9860} & 0.7083 & 0.2787 & 0.9500 & \textbf{0.6374} & \textbf{0.5723} \\
      & 0.1 (Ours) & \underline{0.9164} & \underline{0.9529} & \underline{0.9848} & 0.7208 & 0.2795 & \textbf{0.9580} & \underline{0.6368} & \underline{0.5712} \\
      & 0.25       & 0.9079 & 0.9511 & 0.9828 & \underline{0.7556} & \underline{0.2822} & \underline{0.9480} & 0.6345 & 0.5691 \\
      & 0.5        & 0.8992 & 0.9492 & 0.9792 & \textbf{0.7889} & \textbf{0.2844} & \textbf{0.9580} & 0.6324 & 0.5669 \\
    \bottomrule
  \end{tabular*}
\end{table}

\section{Limitations and Future Works}
\label{limitations}

Although hybrid noise initialization can achieve a balance between long-range consistency and motion diversity, this strategy may not be effective in all scenarios. Due to randomness in the noise initialization process, our methods cannot fully control the differences and connections between subsequently generated noise clips and the first noise clip. Therefore, these noise clips may turn into video clips that are too similar or significantly different video clips after the denoising process, resulting in repetitive or incoherent long videos. Besides, hybrid noise initialization assumes temporal stationarity of video content, so that our method may face challenges when generating videos with abrupt motions or scene cuts. We will leave this as future work and apply methods such as noise searching to address this limitation. Furthermore, our methods may be expanded to multi-prompt video generation to further improve motion diversity in long videos.

\section{Conclusion}
\label{conclusion}

In this paper, we proposed Diff-VF, a general framework to generate high-quality long videos with existing short video diffusion models. We introduced three strategies to address the challenge of keeping long-range consistency, enhancing visual quality and generating rich motion. We first use hybrid noise initialization to control the beginning stage of the diffusion process, achieving a balance between temporal consistency and motion diversity of the video. In the denoising process, we use weighted window sampling and temporal extended sampling to enhance motion smoothness and long-range consistency. For video enhancement, we additionally proposed skip residual guidance to improve fidelity between generated videos and original videos. Experiments show that Diff-VF achieves a favorable balance between temporal consistency and motion diversity compared with previous training-free long-video generation methods. By combining HNI, WWS, and TES, Diff-VF reduces the tendency toward repetitive or static long videos while maintaining competitive temporal coherence and frame-wise quality. Furthermore, we demonstrate that our method can be applied to video diffusion models with 3D VAEs and 3D attention modules, and can generate longer videos with up to $4$ times the original length.

\begin{acks}
This work was supported by the National Natural Science Foundation of China (62471290,62331014) and the Fundamental Research Funds for the Central Universities.

This work was also supported by Shanghai Artificial Intelligence Laboratory.
\end{acks}

\bibliographystyle{ACM-Reference-Format}
\bibliography{main}

\newpage

\appendix

\section{More Qualitative Comparison Results}

\subsection{LaVie}

\begin{figure}[!h]
    \centering
    \includegraphics[width=\linewidth]{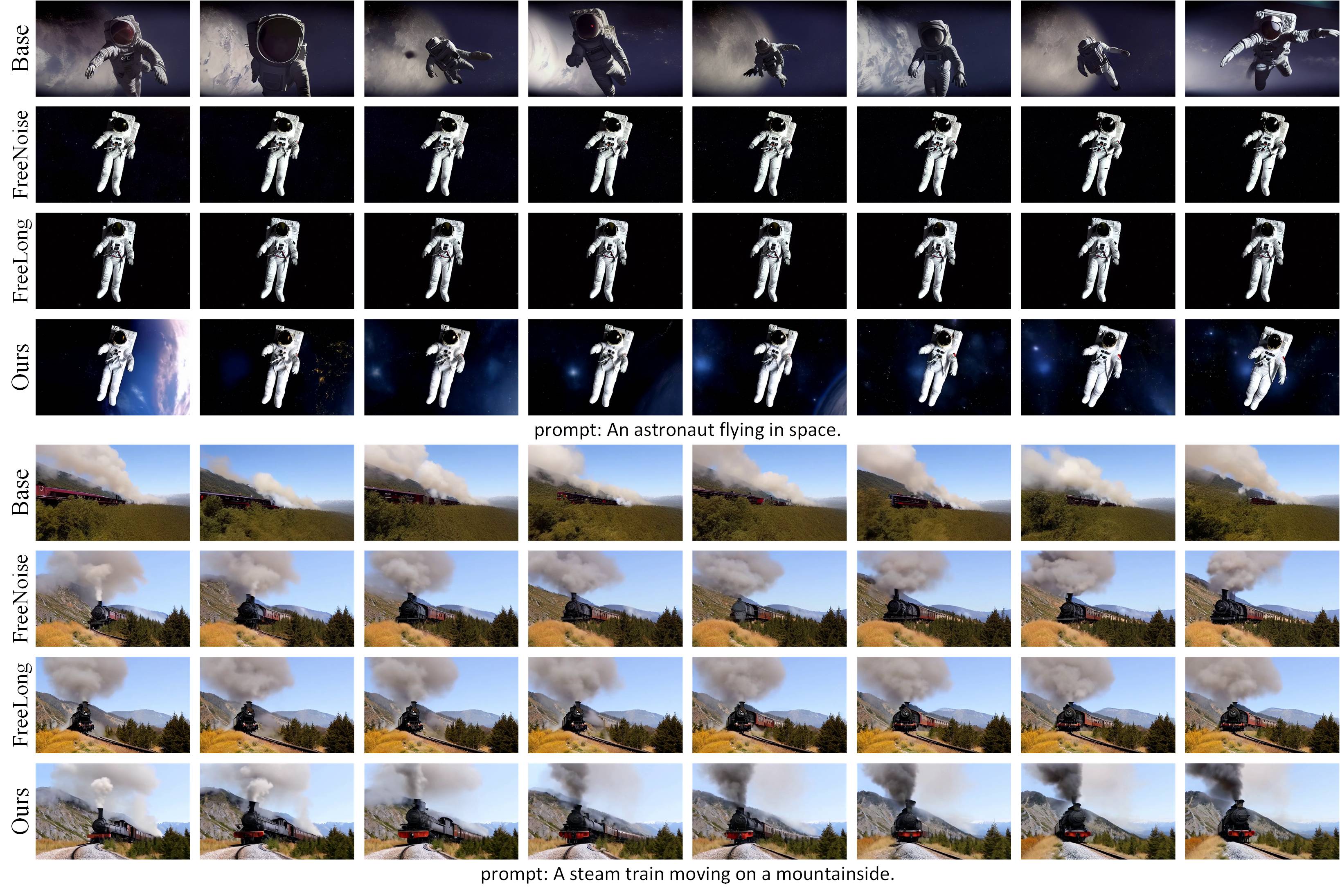}
    \caption{More qualitative comparison based on LaVie. Complete videos corresponding to the visual results in this figure are provided in the supplementary material.}
    \label{lavie_appendix}
\end{figure}

\cref{lavie_appendix} shows more qualitative comparison results based on LaVie \cite{lavie}. Videos generated directly by the base model suffer from poor imaging quality and temporal flickering, and those generated by FreeNoise and FreeLong have limited motions, or are even static. Our method can generate videos that are both temporal consistent and dynamic.

\subsection{HunyuanVideo}

\begin{figure}
    \centering
    \includegraphics[width=\linewidth]{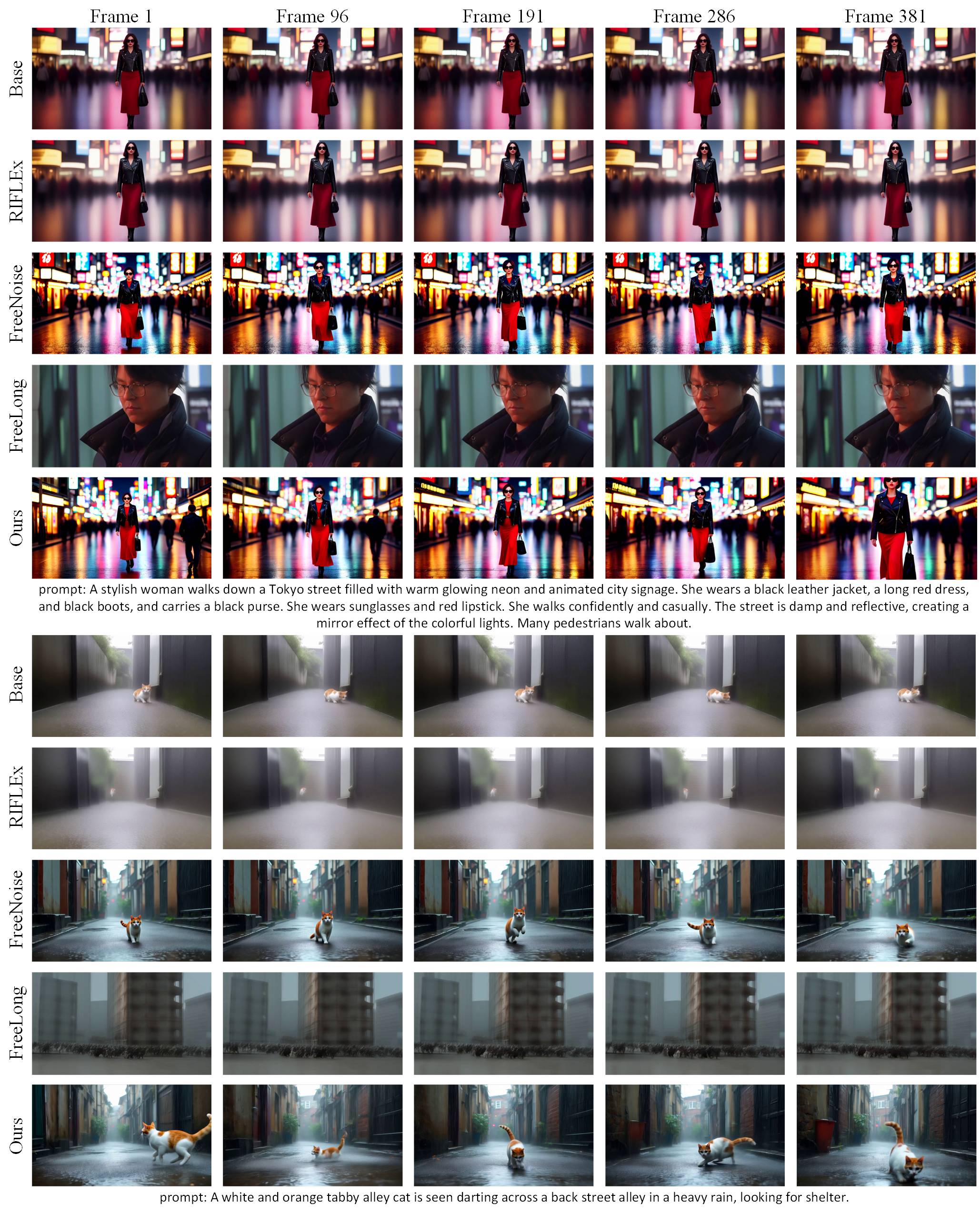}
    \caption{Qualitative comparison based on HunyuanVideo (global view). Complete videos corresponding to the visual results in this figure are provided in the supplementary material.}
    \label{hunyuan_appendix1}
\end{figure}

\begin{figure}
    \centering
    \includegraphics[width=\linewidth]{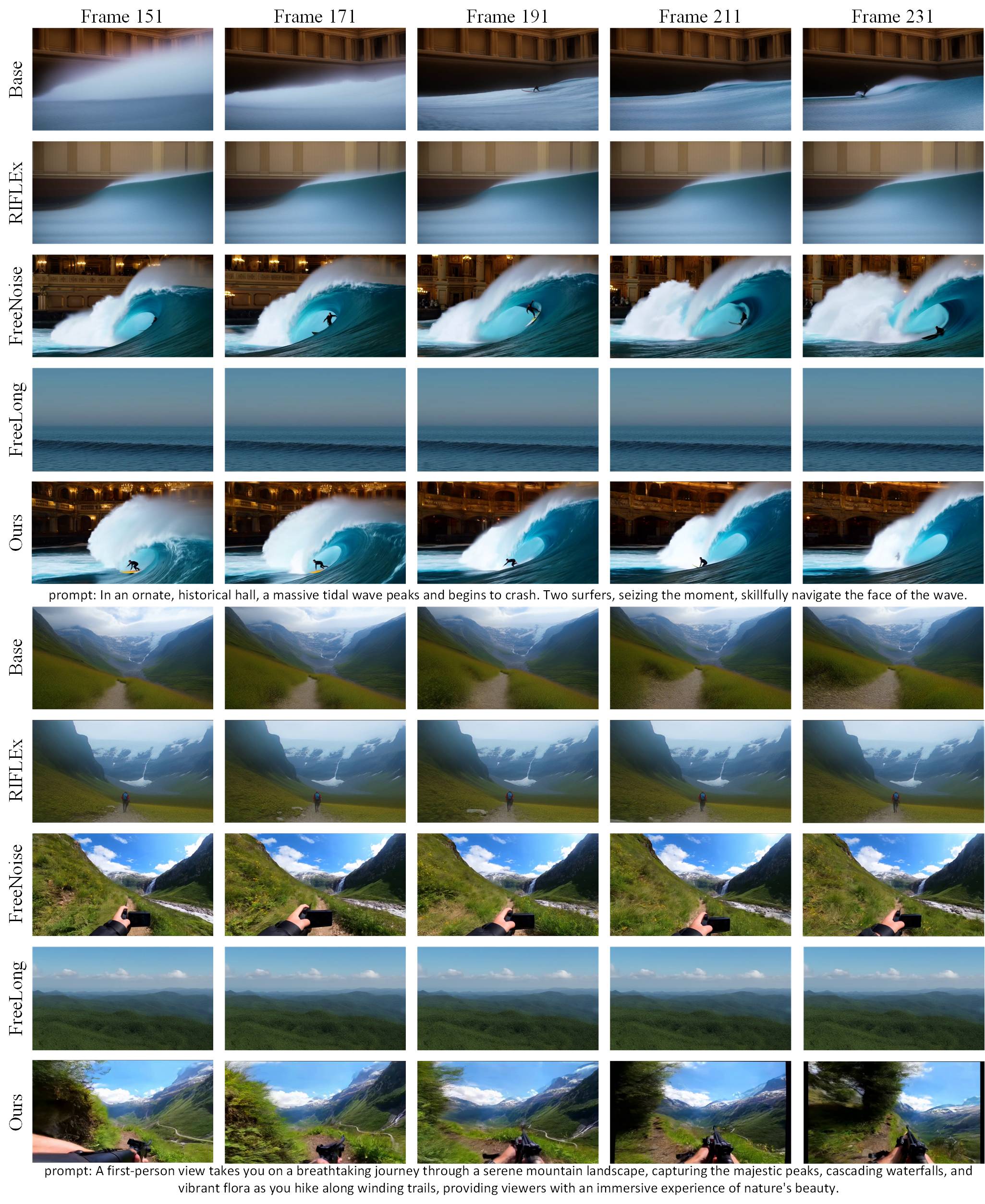}
    \caption{Qualitative comparison based on HunyuanVideo (local view). Complete videos corresponding to the visual results in this figure are provided in the supplementary material.}
    \label{hunyuan_appendix2}
\end{figure}

\cref{hunyuan_appendix1} and \cref{hunyuan_appendix2} present additional qualitative comparison results based on HunyuanVideo \cite{hunyuanvideo}. In \cref{hunyuan_appendix1}, frames are sampled from different temporal locations across the entire video to illustrate long-range consistency, while \cref{hunyuan_appendix2} focuses on frames extracted from short temporal segments to highlight local motion continuity.

Videos generated by the base model, RIFLEx, and FreeLong exhibit noticeable degradation in visual quality. From a global perspective, FreeNoise fails to produce sufficiently pronounced motions that align with the textual prompts. From a local perspective, the generated motions are either overly static (e.g., the subject’s motion in the first example of \cref{hunyuan_appendix2}) or temporally discontinuous (e.g., the background motion in the second example of \cref{hunyuan_appendix2}).

In contrast, our method preserves both temporal consistency and high visual fidelity, while simultaneously generating dynamic and smooth motions that are well aligned with the input text prompts. This demonstrates the effectiveness of our approach in balancing long-range coherence with fine-grained motion continuity.

\clearpage

\clearpage

\section{Extension to other base models}

\begin{figure}[!h]
    \centering
    \includegraphics[width=\linewidth]{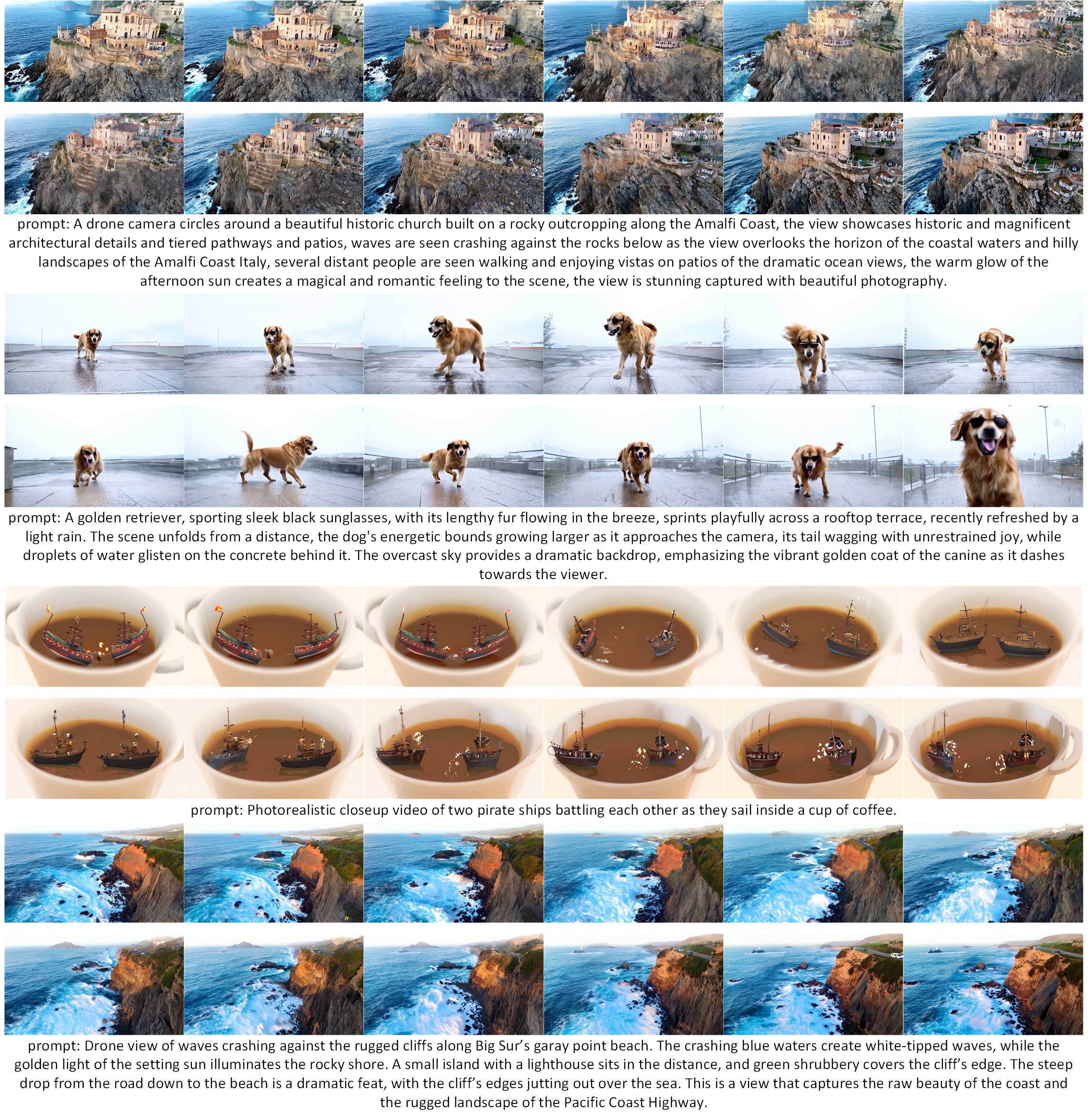}
    \caption{Illustration of long videos generated by our method based on CogVideoX1.5-5B. Complete videos corresponding to the visual results in this figure are provided in the supplementary material.}
    \label{cogvideox_show}
\end{figure}

\begin{figure}[!h]
    \centering
    \includegraphics[width=\linewidth]{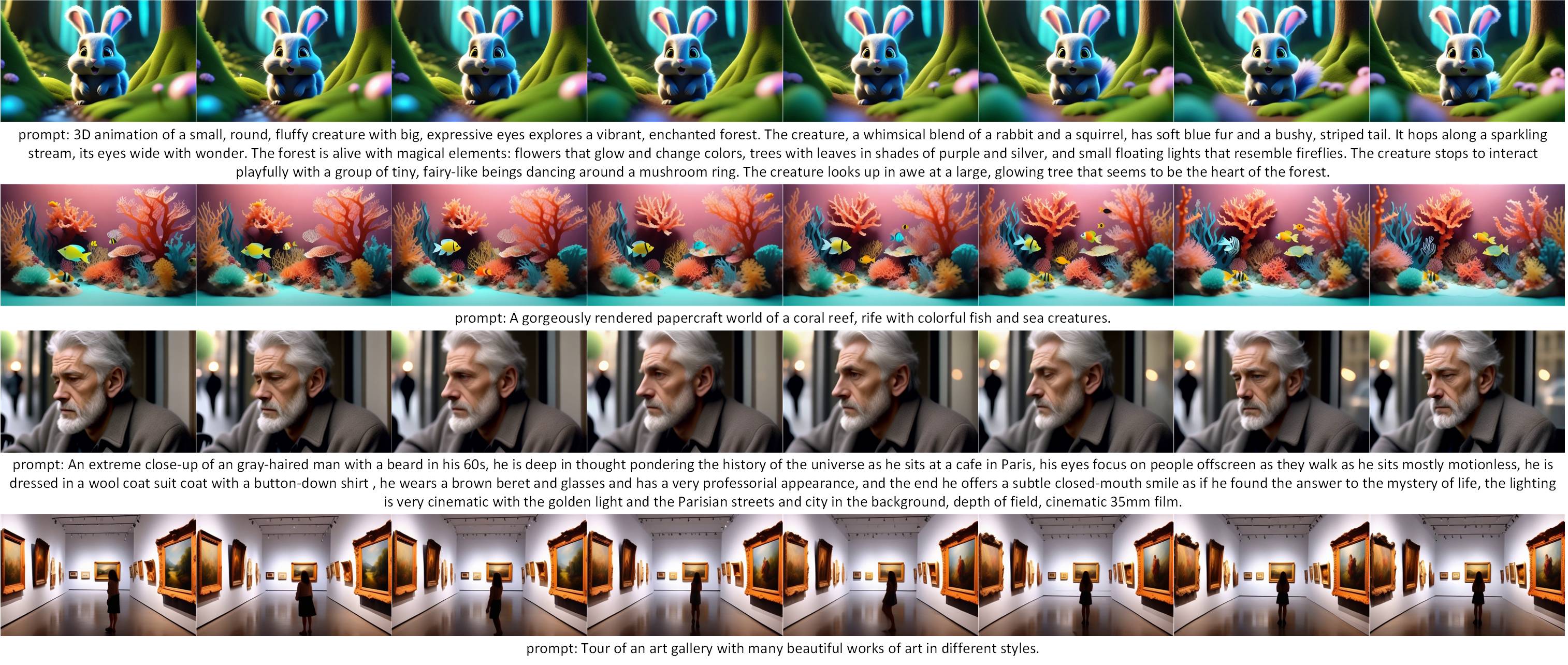}
    \caption{Illustration of long videos generated by our method based on VideoCrafter2. Complete videos corresponding to the visual results in this figure are provided in the supplementary material.}
    \label{vc2_show}
\end{figure}

Besides LaVie and HunyuanVideo, our method can be applied to other base models, such as CogVideoX1.5-5B \cite{cogvideox} (\cref{cogvideox_show}) and VideoCrafter2 \cite{videocrafter2} (\cref{vc2_show}).

\end{document}